\documentclass[english]{article}

\usepackage[preprint,nonatbib]{nips_2018_wider_Nonotice}
\usepackage[T1]{fontenc}
\usepackage[latin9]{inputenc}
\usepackage{color,colortbl}
\usepackage{babel}
\usepackage{verbatim}
\usepackage{url}
\usepackage{amsmath}
\usepackage{amssymb}
\usepackage{graphicx}
\usepackage{setspace}
\usepackage{cancel}
\usepackage{hyperref}

\hypersetup{linkcolor=blue,filecolor=magenta,urlcolor=cyan} 
\usepackage{courier}

\usepackage{amsfonts}

\DeclareMathOperator{\proj}{proj}
\newcommand{\rvline}{\hspace*{-\arraycolsep}\vline\hspace*{-\arraycolsep}}

\newcommand{\bb}[1]{\mathbb{#1}}

\newcommand{\R}{\bb{R}}

\newcommand{\lr}[1]{\left(#1\right)}

\newtheorem{lemma}{Lemma}

\newtheorem{thm}{Theorem}

\newcommand{\ka}{\kappa}

\begin{document}

\title{Polysemanticity and Capacity in Neural Networks}

\author{Adam Scherlis$^{1}$, Kshitij Sachan$^{1}$, Adam S. Jermyn$^{2}$, Joe Benton, Buck Shlegeris$^{1}$\\\\
$^{1}$Redwood Research\\
$^{2}$Flatiron Institute}

\maketitle
\begin{abstract}
Individual neurons in neural networks often represent a mixture of unrelated features. This phenomenon, called polysemanticity, can make interpreting neural networks more difficult and so we aim to understand its causes.
We propose doing so through the lens of  feature \emph{capacity}, which is the fractional dimension each feature consumes in the embedding space.
We show that in a toy model the optimal capacity allocation tends to monosemantically represent the most important features, polysemantically represent less important features (in proportion to their impact on the loss), and entirely ignore the least important features. Polysemanticity is more prevalent when the inputs have higher kurtosis or sparsity and more prevalent in some architectures than others.
Given an optimal allocation of capacity, we go on to study the geometry of the embedding space.
We find a block-semi-orthogonal structure, with differing block sizes in different models, highlighting the impact of model architecture on the interpretability of its neurons.

\end{abstract}

\section{Introduction}
Individual neurons in neural networks often represent multiple unrelated features in the input~\cite{olah2017feature,olah2020zoom}.
This phenomenon is known as polysemanticity, and makes it more difficult to interpret neural networks~\cite{olah2020zoom}.
While "feature" is a somewhat fuzzy concept~\cite{XYZ}, there are at least some cases where we ``know it when we see it''.
For example, when the input features are independent random variables that do not interact in the data-generating process, neurons that represent combinations of these input features can be confidently called polysemantic.
In this work we explore how loss functions incentivize polysemanticity in this setting, and the structure of the learned solutions.

\begin{figure*}
    \centering
    \includegraphics[width=0.6\textwidth]{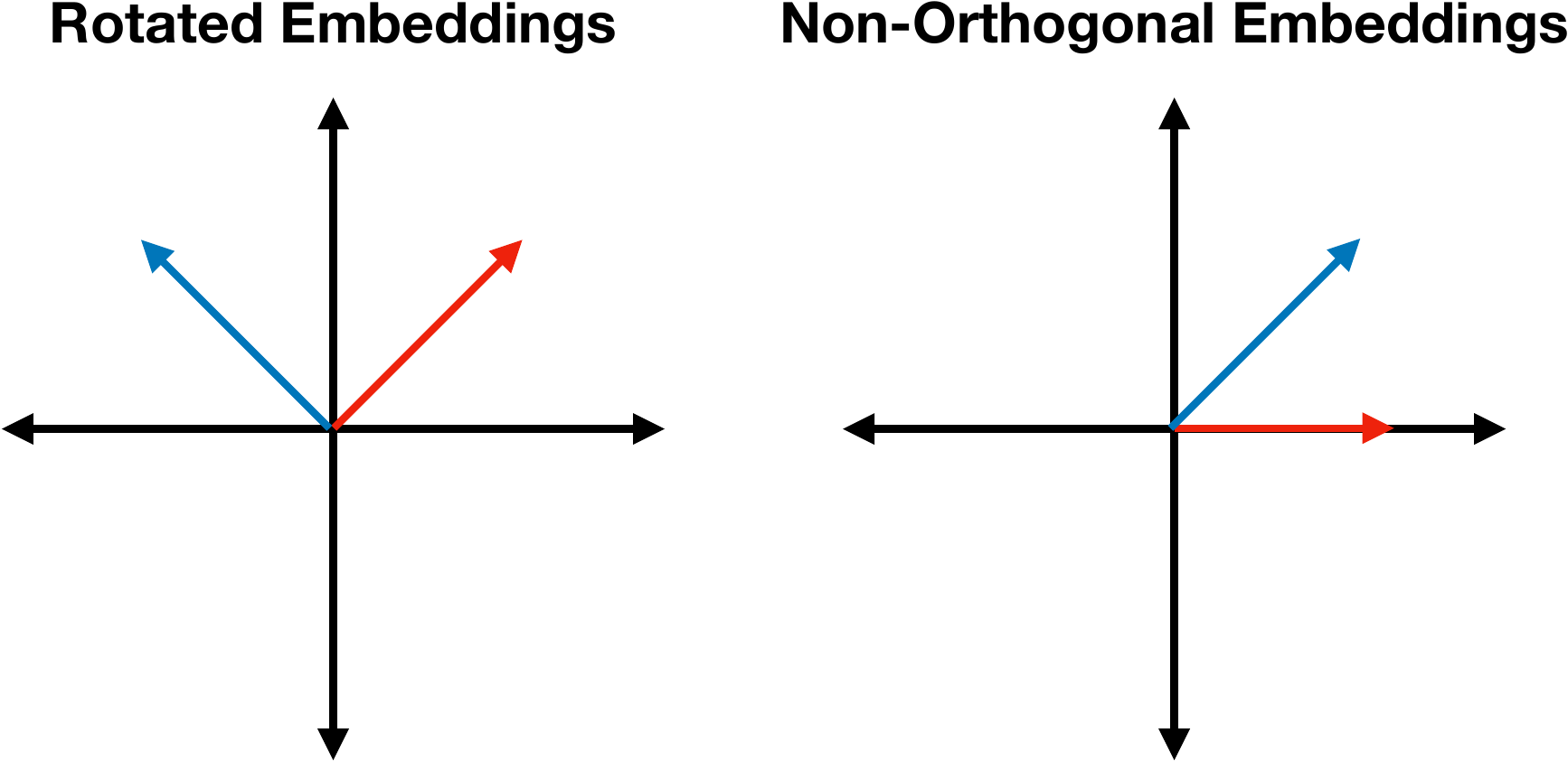}
    \caption{Feature embedding vectors are shown in two dimensions. The neuron basis corresponds to the coordinate axes. Left: rotated embeddings. Right: non-orthogonal embeddings. In both cases the result is polysemanticity because each neuron receives some input when either feature is present.}
    \label{fig:polysemaniticy vs superposition}
\end{figure*}

Fittingly, there are multiple ways that polysemanticity can manifest.
Here we focus on one form that seems particularly fundamental, namely superposition~\cite{XYZ}.
Suppose we have a linear layer that embeds features which then pass through a layer with a nonlinear activation function.
The feature embedding vectors might not be orthogonal, in which case multiple neurons (nonlinear units) are involved in representing each feature.
When there are at least as many features as neurons this means that some neurons represent multiple features, and so are polysemantic (Figure \ref{fig:polysemaniticy vs superposition}, right).
There are other causes of polysemanticity, e.g. feature embedding vectors could be rotated relative to the neuron basis (Figure \ref{fig:polysemaniticy vs superposition}, left), but we do not study these in this work.

Here we build on the work of \cite{XYZ}, who studied polysemanticity in the context of toy models of autoencoders.
They found that models can support both monosemantic and polysemantic neurons, that polysemantic neurons can perform certain kinds of computations, and that the embedding vectors of features often formed repeating motifs of a few features symmetrically embedded in a low-dimensional subspace.
Moreover, in their models  they found distinct ``phases'' where superposition was either significant or completely absent.
Sparser inputs resulted in more superposition. Features with similar importance were more likely to be in superposition.
This reflects an abundance of unexpected structure, and gives new handles on the phenomenon of polysemanticity.

\begin{figure*}
    \centering
    \includegraphics[width=0.6\textwidth]{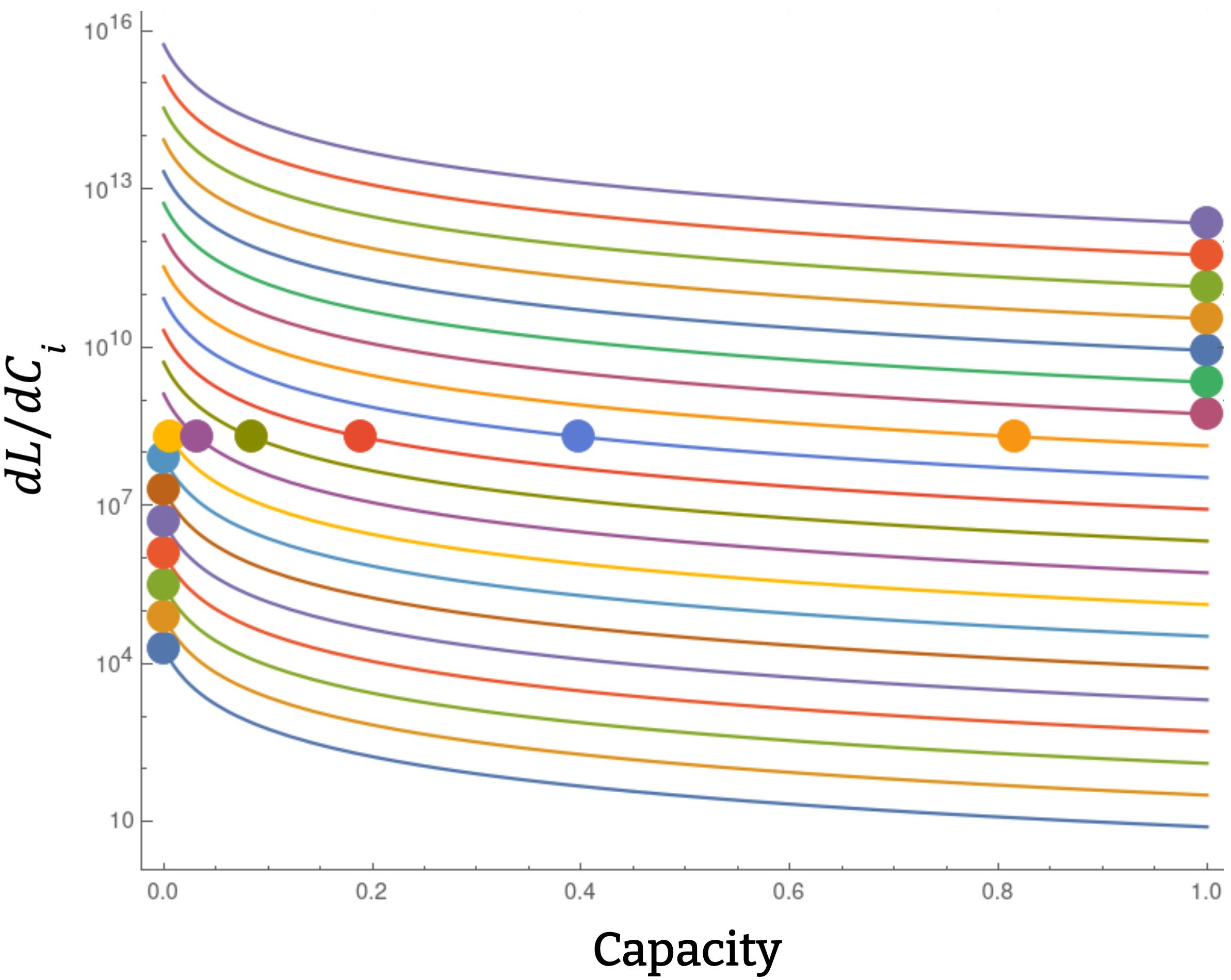}
    \caption{The marginal loss reduction $-\partial L/\partial C_i$ is shown for several features as a function of feature capacity in our toy model. Circles represent the optimal capacity allocation for a particular total embedding dimension. Colors vary to make individual curves more distinguishable.}
    \label{fig:deriv-curves}
\end{figure*}

We study these phenomena through the lens of \emph{capacity}, or the fraction of an embedding dimension allocated to each feature (Section~\ref{sec:capacity}, also termed ``dimensionality'' by \cite{XYZ}).
This ranges from 0-1 for each feature, and the total capacity across all features is bounded by the dimension of the embedding space.
Because the model has a limited number of neurons and so a limited number of embedding dimensions, there is a trade-off between representing different features.
We find that the capacity constraint on individual features (0-1) means that many features are either ignored altogether (not embedded) or else allocated a full dimension orthogonal to all the other features in the embedding space, depending on the relative importance of each feature to the loss.
Features are represented polysemantically only when the marginal loss reduction of assigning more capacity to each is equal (Figure \ref{fig:deriv-curves}).
This neatly explains the sharp ``pinning'' of features to either 0 or 1 capacity noted by \cite{XYZ}, and gives us a framework for understanding the circumstances under which features are represented polysemantically.

\begin{figure*}
    \centering
    \includegraphics[width=0.99\textwidth]{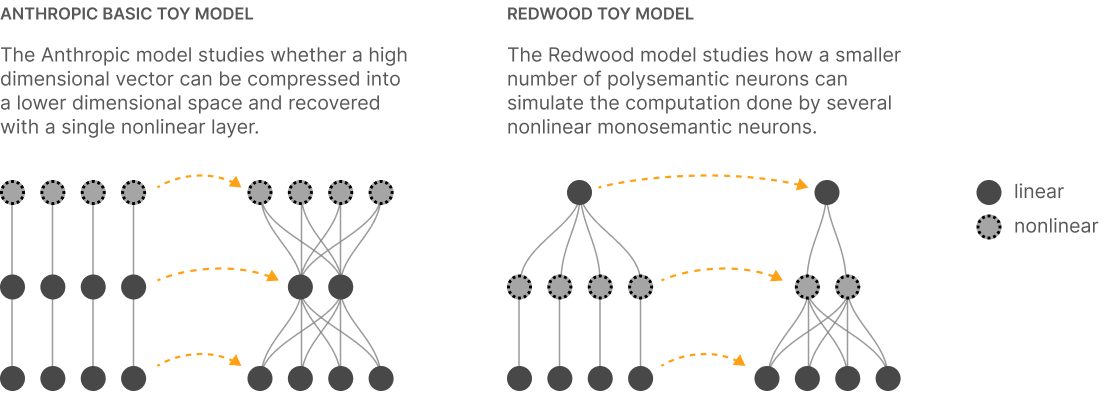}
    \caption{Comparison between the Anthropic toy model of~\cite{XYZ} (left) and our toy model (right). Model inputs are at the bottom of the diagram and outputs are at the top. The key difference is that the Anthropic model studies the compression and recovery of high-dimensional vectors, while ours examines how a smaller number of polysemantic neurons can simulate the computation done by a larger number of monosemantic ones. Figure kindly provided by Chris Olah.}
    \label{fig:schema2}
\end{figure*}

To explore capacity allocation in a concrete model, we instantiate our theory for a one-layer model with quadratic activations (Section~\ref{sec:quadratic}). Our model differs from the Anthropic toy model in that ours uses a different activation function to make the math more tractable, and, more importantly, ours is focused on polysemantic computation rather than data compression.
We contrast these toy models in Figure~\ref{fig:schema2}.

\begin{figure*}
    \centering
    \includegraphics[width=0.98\textwidth]{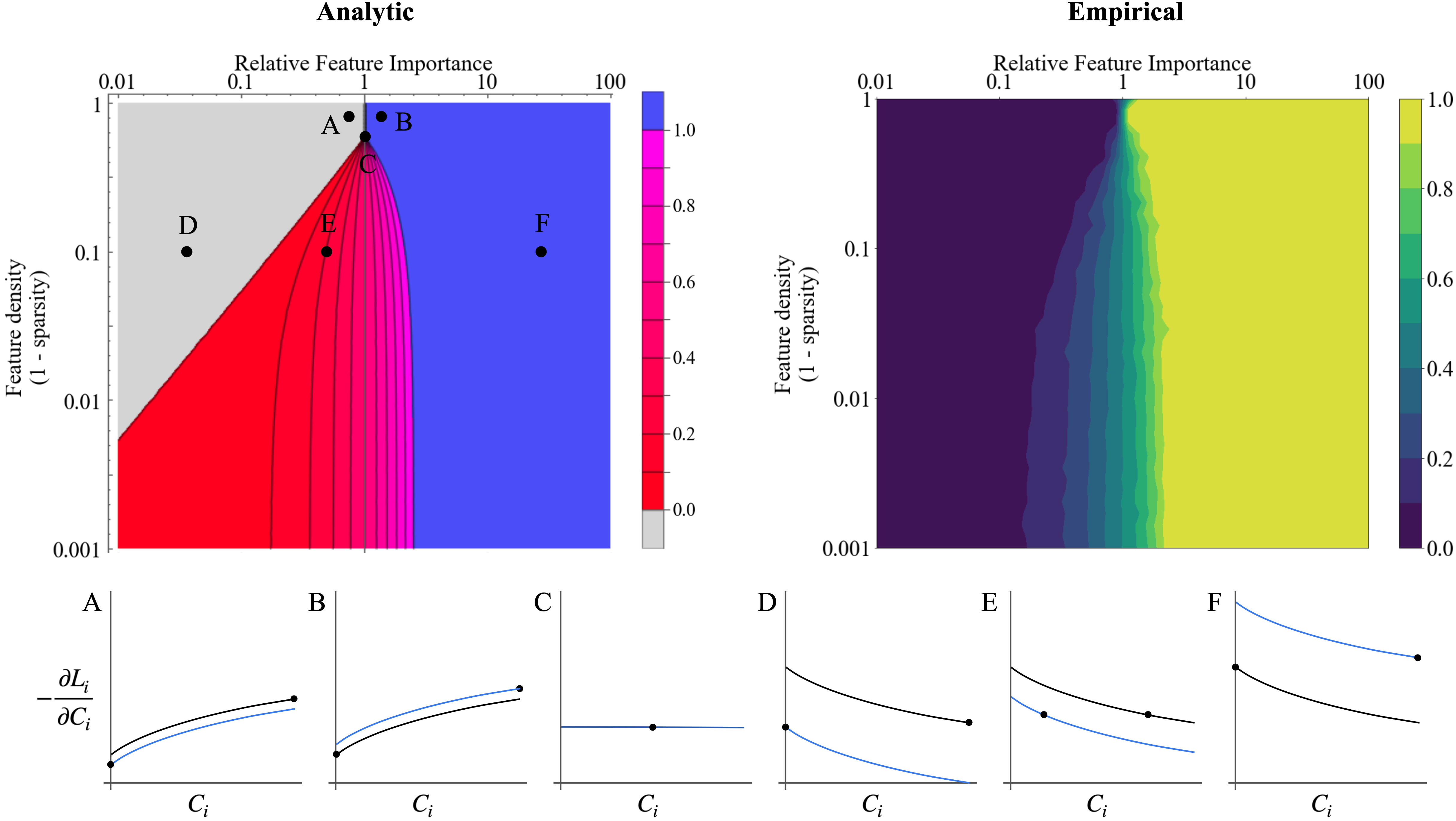}
    \caption{Upper: Analytical and empirical phase diagrams for our toy model with 6 features and 3 neurons. In both panels one feature has a different importance from the rest, and colors show the resulting capacity allocation for that feature as a function of sparsity and relative importance. Lower: Plots of marginal loss reduction $\partial L/\partial C_i$ as a function of feature capacity for each labelled point in the analytical phase diagram. The blue curve represents the feature with varied importance and the black one represents the constant important feature. Black dots are optimal allocations of capacity.}
    \label{fig:phase-analytic}
\end{figure*}

For our toy model we can analytically determine the capacity allocation as a function of feature sparsity and importance (i.e. weight in the loss), and so construct a ``phase diagram'' (Figure \ref{fig:phase-analytic}).
While the details of our phase diagram differ from those of \cite{XYZ}, reflecting our different toy model, there are three qualitative features that are in good agreement.
First, when a feature is much more important than the rest, it is always represented fully with its own embedding dimension.
Second, when a feature is much less important than the rest, it is ignored entirely.
Finally, in a sparsity-dependent intermediate region features are partially represented, sharing embedding dimensions.
In addition, this confirms our theoretical expectation that capacity is allocated according to how much each feature matters to the loss (a mixture of importance and sparsity) and that it is often allocated to fully ignore some features while fully representing others.
We supplement this with empirical results for a variety of activation functions showing that the phase diagram predicts the behavior of a broad family of 2-layer models.

We then turn to study the geometry of the embedding space (Section~\ref{sec:matrices}).
When embedding matrices fully utilize the available capacity we call them ``efficient''.
We find that every efficient embedding matrix has a block-semi-orthogonal structure, with features partitioned into different blocks.
When multiple features in a block are present they interfere with each other, causing spurious correlations in the output and hence greater loss.
Features do not, however, interfere across blocks.

\begin{figure*}
    \centering
    \includegraphics[width=0.9\textwidth]{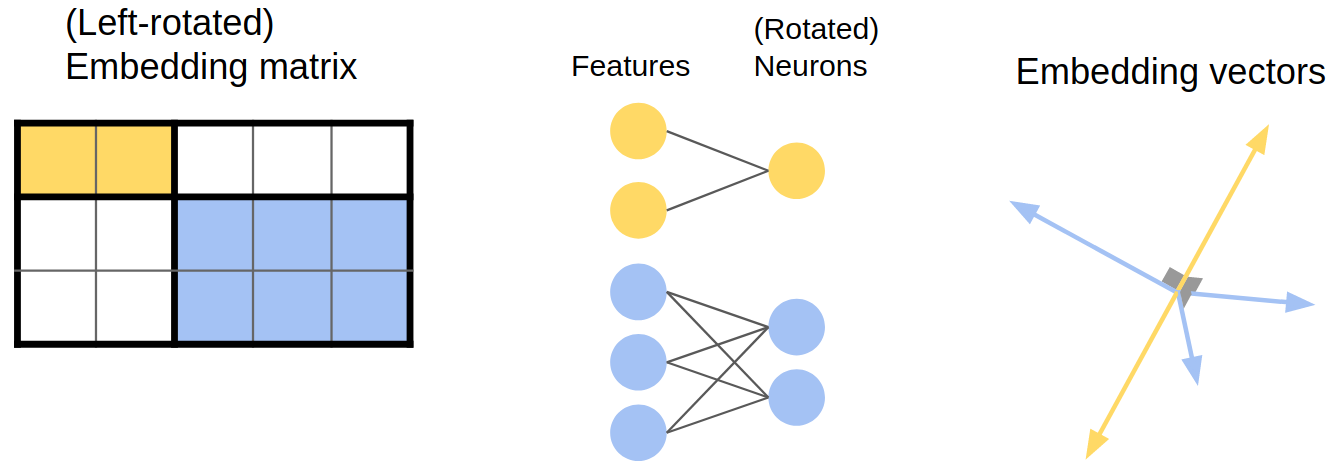}
    \caption{Left: An embedding matrix with two blocks. Center: The relationship between features and (principal-component-aligned) neurons for this matrix. Right: Embedding vector geometry for this matrix.}
    \label{fig:embedding-geometry}
\end{figure*}

The blocks in efficient matrices correspond to the polytope structure \cite{XYZ} found, with small blocks corresponding to features embedded as regular polytopes and large blocks corresponding to less-ordered structures.
Large- and small-block arrangements come with different advantages.
With large blocks there is significant freedom to allocate capacity across features, whereas with small blocks there is the additional constraint that the capacity of each block be an integer and that the block capacities add up to the total capacity.
On the other hand, with small blocks the lengths of embedding vectors can be chosen more freely because blocks can be scaled independently of each other without affecting the capacity allocation.

In our quadratic model the embedding matrices in our toy model always have one large block, which is correspondingly less structured.
We expect that differences in architecture can lead to different sizes of blocks, which could provide a way to control the extent of polysemanticity in models, alongside other approaches such as changing the activation function~\cite{elhage2022solu}.

\section{Capacity and Superposition}\label{sec:capacity}

\subsection{Definitions}

Suppose we have a model composed of stacks of linear layers with nonlinear activation functions. In each layer, the model applies a linear transform to the input vector $x$ to produce an embedding vector $e$, and then performs an element-wise non-linear calculation on those embeddings to produce the non-linear activation vector $h$. For instance, we might have

\begin{align}
e&\equiv W\cdot x\\
h&\equiv \mathrm{ReLU}(e)
\end{align}

with $W\in \mathbb{R}^{d \times p}$, $x \in \mathbb{R}^p$, $e,h\in \mathbb{R}^d$. We associate each dimension of the input vector $x$ with a feature, and we call each dimension of the non-linear layer a neuron. 

For simplicity, in the rest of this paper we work with a one-layer model, but our capacity definition should be valid for any layer in a multi-layer model.

When a model represents a feature in the input space, it is convenient to think that it expends some capacity to do so. Our intuition here is that as we ask a model to represent more and more features we eventually exhaust its ability to do so, resulting in features interfering. We study the superposition phenomena by asking the question: "How do models allocate limited representation capacity to input features?" In what follows we assume that each input feature is assigned a unique dimension in the input space (e.g. feature $i$ is input dimension $i$), and capacity we define below. 

Let $W_{\cdot,i} \in \mathbb{R}^d$ be the embedding vector for feature $i$. The capacity allocated to feature i is
\begin{align}
C_i=\frac{(W_{\cdot,i}\cdot W_{\cdot,i})^2}{\sum_j (W_{\cdot,i}\cdot W_{\cdot,j})^2}
\label{eq:capacity}
\end{align}

We can think of $C_i$ as ``the fraction of a dimension'' allocated to feature $i$ (\cite{XYZ})\footnote{We can also interpret $C_i$ as the squared correlation coefficient between $x_i$ and $(W^TWx)_i$ -- see Appendix \ref{app:rho2}.}. The numerator measures the size of the embedding and the denominator tracks the interference from other features. By this definition, $C_i$ is bounded between 0 and 1. In the case $W_{\cdot,i}=0$, where this expression is undefined, we set $C_i=0$.\footnote{This is the limit of the expression from almost all directions, so in practice $W_{\cdot,i}\approx 0$ implies $C_i\approx 0$.} 

We define the total model capacity to be $C=\sum_i C_i$ (in a multi-layer model, this would be a single layer-pair's capacity). This is bounded between 1 and the embedding dimension $D$ (see Appendix \ref{app:cap-constraint} for a proof of the upper bound).

\begin{figure*}
    \centering
    \includegraphics[width=0.96\textwidth]{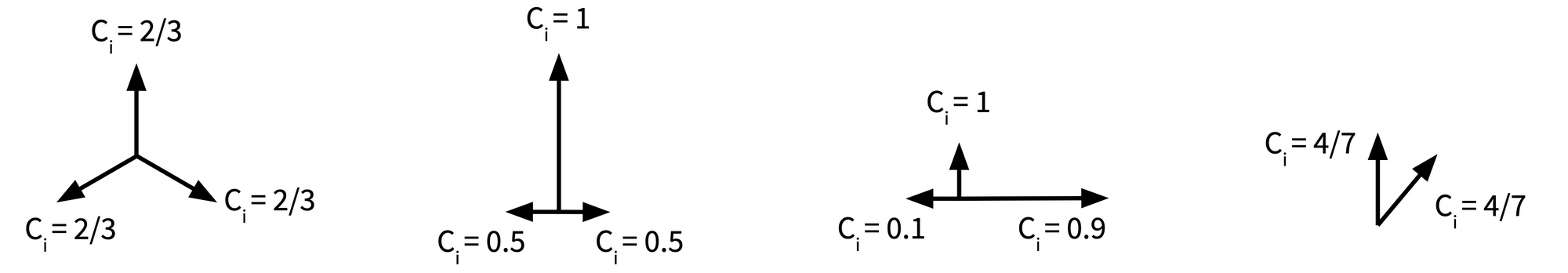}
    \caption{Example capacity allocations for different embeddings.}
    \label{fig:my_label}
\end{figure*}

Note that a set of capacities does not uniquely specify a weight matrix. For example, capacity is invariant to the overall scaling and rotation of $W$. In what follows it will be useful to have a full parameterization of $W$ that includes $C_i$, so we define $S$ to be a set of additional parameters that uniquely specify a weight matrix $W$ given its capacities $C_1,\ldots,C_N$. We can then parameterize the loss using  $(C_1,\ldots,C_N,S)$ rather than $W$.

\subsection{Loss Minimization}\label{sec:loss_min}

We are interested in how loss minimization allocates capacity among different features.
Because the capacity of each feature lies in $[0,1]$ and there is also a constraint on the total capacity of a model, this is a constrained optimization problem:
$$
\begin{aligned}
\min_{C_{1:n},S} \quad & L(C_{1:n},S)\\
\textrm{s.t.} \quad & 0 \leq C_i \leq 1\\
  &1\le \sum_i C_i \leq D    \\
\end{aligned}
$$

In brief, we minimize over $S$ first to find the $S^*(C_i)$ minimizing the loss\footnote{There are multiple ways to define $S$, which give different $(\partial L / \partial C_i)$ holding "$S$" constant. Minimizing over $S$ removes this ambiguity.}.
We then minimize over $C_i$
Intuitively capacity should always be a good thing, so we are interested in situations where the loss function $L$ is monotonically decreasing in $C_i$. In this case, the bound on the total capacity $\sum_i C_i \le D$ will be saturated, and we can replace this constraint with a Lagrange multiplier:
\begin{align}
\label{eq:Lagrange}
\min_{C_{1:n},\lambda} \quad & L(C_{1:n}) + \lambda \left(\sum_i C_i - D\right)\\
\textrm{s.t.} \quad & 0 \leq C_i \leq 1\nonumber
\end{align}

Because we have independent constraints on each $C_i$, for each $i$ the minimum occurs either at a boundary ($C_i=0$ or $1$) or in between at a point where
\begin{align}
    \frac{\partial}{\partial C_i} \left( L(C_{1:n}) + \lambda \left(\sum_i C_i - D\right)\right)=0
\end{align}

The derivative in the three cases will obey
\begin{align}
    \frac{\partial L(C_{1:n})}{\partial C_{i}} &\ge -\lambda \quad  (C_i = 0)\\
    \frac{\partial L(C_{1:n})}{\partial C_{i}} &= -\lambda \quad (0 < C_i < 1)\\
    \frac{\partial L(C_{1:n})}{\partial C_{i}} &\le -\lambda \quad (C_i = 1)
\end{align}

In other words, at the minimum of the loss, capacity is allocated to each feature until the marginal value (decrease in loss) from additional capacity is at a constant threshold -- unless the feature is so important that the marginal value is above threshold even when fully represented, or so unimportant that the marginal value is below threshold even when the feature is ignored.

Note that none of this derivation depended on the precise definition of capacity. We only had to assume that (1) allocating more capacity to a feature never makes the loss greater, (2) all capacity allocations satisfying the feature-level and total constraints are realizable by some embedding matrix.
Thus we could have used other definitions, and indeed in Appendix~\ref{app:alt-capacity} we explore one such alternative.

We can visualize this in terms of the graph of $\partial L/\partial C_i$ as a function of $C_i$, holding $C_j$ constant for $j\ne i$.
For instance, in Figure~\ref{fig:deriv-curves} there are diminishing returns: $\partial^2 L/\partial C_i^2 > 0$, so that each additional unit of capacity makes the next one less useful. If $\partial L/\partial C_i$ crosses $-\lambda$, the value of $C_i$ where it crosses will be the optimal allocation. Features only get 0 or 1 capacity if the entire $\partial L/\partial C_i$ curve is above or below $-\lambda$ (respectively).

By contrast, in the case of increasing returns ($\partial^2 L/\partial C_i^2 < 0$), every feature will have $C_i=0$ or $C_i=1$ and superposition will not occur at all.
As we show below for several different toy models, diminishing returns occur for inputs of high sparsity or kurtosis, and accelerating return for inputs of low sparsity or kurtosis.

\section{Quadratic model}\label{sec:quadratic}

We now instantiate our theory for a two-layer model with quadratic activations.
This choice of activation function allows us to study this model analytically in some detail.

We define our model in Section~\ref{sec:model}.
In Section~\ref{subsec:interpretation} we analyze the loss function for the case of independent random input features.
We are able to write the expected loss entirely in terms of the embedding lengths and the variance and kurtosis of feature strengths.

In Section~\ref{subsec:dLdC} we demonstrate a surprising feature of this quadratic model: the loss function can be written in terms of the feature capacity and the embedding lengths.
This allows us to analytically explore the marginal benefit of allocating more capacity to each feature.

We then go on to solve for the optimum allocation of capacity across features (Section~\ref{subsec:solve}) and relate this to our results from Section~\ref{sec:capacity}).

\subsection{The model}\label{sec:model}
We study a regression task, with ground-truth data $x,y(x)$ and modeled data $x,\tilde y(x)$.
The ground truth data is generated by a random vector of IID variables, $x$, and a set of ``importance" coefficients $v_i$:
\begin{align}
    y&=\sum_i v_i x_i^2.
\end{align}
We think of $v_i$ as the importance of feature $i$, and highlight that $y$ here is a scalar. The model $\tilde y(x)$ is parameterized as
\begin{align}
\tilde{y}&=\sum_i (W_{i, \cdot} \cdot x)^2+b
\end{align}
where $b$ is a bias and $W$ is a rectangular embedding matrix.
Like $y$, $\tilde{y}$ is a scalar.
Note that the ground-truth and modeled data are both of the form $x^T A x$ for a matrix $A$. For the ground truth, this is a full-rank diagonal\footnote{In general, we could consider a ground truth with an arbitrary full-rank $A$ instead of a diagonal matrix. Forcing $A$ to be diagonal is then equivalent to choosing to always write $x$ in the principal-component (SVD) basis of $A$. This gives a slightly stronger motivation for our choice to interpret the components of $x$ as separate ``features": we are really choosing features based on the principal axes of $A$.} matrix; for our model it is limited in rank by the shape of $W$. Therefore, perfect loss is not possible in general.
 
Our loss function is squared error
\begin{align}
L(x) = (\tilde y(x) - y(x))^2.
\end{align}

\subsection{Interpreting the loss function}\label{subsec:interpretation}

We now assume that $\mathbb{E}[x_i]=0$ and $\text{Cov}[x_i,x_j]=\delta_{ij}$, but do not impose any additional restrictions on the distribution of $\{x_i\}$.
With this, we find (Appendix~\ref{appen:loss})
\begin{equation}
\mathbb{E}[L]= (\mathbb{E}[x_i^4]-1)\sum_i (||W_{\cdot,i}||^2-v_i)^2 +2\sum_{i \neq j}(W_{\cdot,i} \cdot W_{\cdot,j})^2 
\end{equation}

The loss is a weighted sum of two components.
The first term captures true correlations in the data, because $v_i \propto \text{Cov}[y,x_i^2]$, while the second represents ``hallucinated correlations'' which are not present in the data ($W_{\cdot,i} \cdot W_{\cdot,j} \propto \text{Cov}[\tilde{y},x_ix_j]$) but which the model produces anyway (Appendix~\ref{appen:covar}).
When the fourth moment is large (for example due to sparsity, see Appendix~\ref{sec:misc}) the hallucinated term becomes less important and we get polysemanticity/superposition.

More generally, when a model is incapable of representing the precise correlations in the data there is often a tradeoff, where the model can improve its representation of the true correlations at the cost of incorporating hallucinated ones.
This will often be beneficial to the overall loss, but can end up spreading the task of representing a single feature across multiple neurons.
If the model has a limited number of neurons this can result in polysemanticity.

\subsection{Capacity}
\label{subsec:dLdC}

We would like to write the expected loss in terms of the capacity of each feature.
Recalling the definition of capacity, the expected loss can be written as
\begin{align}
\mathcal{L}(\vec C, \vec n) = (k-1)\sum_i (n_i - v_i)^2 - 2\sum_i n_i^2 + 2\sum_i \frac{n_i^2}{C_i},
\end{align}
where $\mathcal{L} \equiv \mathbb{E}[L]$ is the expected loss, $k \equiv \mathbb{E}[x_i^4]$ is the kurtosis, and $n_i \equiv ||W_{\cdot,i}||^2$ is the squared embedding length.
This form is surprising because it says that the loss is composed of (1) a term that depends only on the embedding lengths and (2) a sum of terms that each depend on the capacity of just one feature.
In particular, what is surprising is that the effects of interference between features are entirely captured by the feature capacity, so the second term can be interpreted as the loss due to interference.

Perhaps more striking, the partial derivative of loss with respect to capacity is just
\begin{equation}
\frac{\partial \mathcal{L}}{\partial C_i}=-2\frac{n_i^2}{C_i^2}.
\end{equation}
That is, the marginal benefit of capacity for feature $i$ depends only on the embedding length and capacity of that feature and not at all on any other aspect of the model.
In what follows we make extensive use of this property to find the optimal capacity allocation.

\subsection{Optimal capacity allocation}\label{subsec:solve}

We now minimize the loss to find (Appendix~\ref{appen:minimize}):
\begin{align}
    C_i &= \max\left(\min\left(\frac{k-1}{k-3}\frac{v_i}{\lambda} - \frac2{k-3}, 1\right), 0\right)\\
    n_i &= \begin{cases}
        0 &\quad C_i=0\\
        \frac{k-1}{k-3}v_i-\frac{2\lambda}{k-3} &\quad 0< C_i < 1\\
        v_i&\quad C_i=1
    \end{cases}
\end{align}
Here $\lambda$ is a Lagrange multiplier determined by the condition that $\sum_i C_i = D$.
In Appendix~\ref{appen:minimize} we also verify that this solution is indeed feasible (i.e. there is an embedding matrix producing the above $C_i$ and $n_i$).

For kurtosis $k>3$, features will be ignored if
\begin{align}
\label{eq:ignore}
\frac{k-1}2v_i<\lambda
\end{align}
and fully represented if
\begin{align}
\label{eq:full}
v_i > \lambda
\end{align}

If $k\le 3$, features will always either be ignored or fully represented  because the loss is concave-down (or flat) in $C_i$.
Hence the $D$ features with highest $v_i$ will be represented.

\subsection{Phase diagrams}

We now calculate phase diagrams for capacity allocation.

\subsubsection{Phase diagram boundaries}

First, we solve for $\lambda$ by inserting the set of $v_i$ into equation~\eqref{eq:Ci_alloc}:
$$
C_i = \max\left(\min\left(\frac{k-1}{k-3}\frac{v_i}{\lambda} - \frac2{k-3}, 1\right), 0\right).
$$
We then sum over $i$ and set this equal to $D$, allowing us to solve for $\lambda$.

Features will be polysemantic if (equations~\ref{eq:ignore} and~\ref{eq:full})
$$
v_i<\lambda < \frac{k-1}2v_i
$$
or equivalently
$$
\frac2{k-1}\lambda < v_i < \lambda
$$
Note that for $k<3$, this is impossible, so features will never be partially represented (polysemantic).

\subsubsection{Most-importances-equal case}

We now consider the case with $N$ features, $D<N$ dimensions, and
\begin{align}
v_i&=V\\
v_i&=1&i\ne1
\end{align}
When all features are partially represented we can use the constraint $\sum_iC_i=D$ to solve for $\lambda$ and find
$$
\lambda=\frac{(k-1)(N-1+V)}{(k-3)D+2N}.
$$
Feature \#1 therefore has capacity
$$
C_1=\frac{(k-3)D+2N}{(k-3)(N-1+V)}V-\frac{2}{k-3}.
$$
As a sanity check, this is $D/N$ when $V=1$.

Feature \#1 becomes fully represented $(C_1=1)$ when
$$
V\ge \frac{(k-1)(N-1)}{(k-3)(D-1)+2(N-1)}
$$
When $k$ is large, this approaches $(N-1)/(D-1)$.
When $V\ge(N-1)/(D-1)$, feature \#1 is fully represented regardless of $k$. 

Feature \#1 is ignored $(C_1=0)$ when

$$
V\le \frac{2(N-1)}{(k-3)D+2(N-1)}
$$

When $k$ is large, this goes to zero like $1/k$.

For $k=3$, both bounds are at $V=1$ and the feature instantly changes from being ignored to fully represented when $V$ crosses 1.

\subsection{Phase Diagram Intuition}
Figure \ref{fig:phase-analytic} shows the analytic phase diagram and the corresponding empirical results.
The lower panel shows the marginal loss benefit ($-\partial L / \partial C_i$) as a function of feature capacity for each labelled point in the analytical phase diagram.

We first focus on relative feature importance.
In our model an increase in relative feature importance corresponds to shifting $\partial L/\partial C_i$ up by a constant factor. Going from point D to point F, the blue feature increases in relative importance, and the corresponding marginal loss curve is shifted up. At point D the blue feature is too unimportant to be represented, at point E the blue feature is similar enough in importance to be allocated a fractional dimension with equal marginal loss benefit to the black reference feature, and at point F the blue feature is so much more important than the black feature that it consumes all available capacity.

We next turn to sparsity (kurtosis).
As sparsity increases, the marginal loss curves go from convex (A) to flat (C), and eventually to concave (E). Thus, there are diminishing returns to representing a feature when it is sparse and increasing returns when it is dense. When features are dense, the curves are concave, so the feature with larger marginal loss benefit is purely represented and the other feature is ignored. This explains the jump from ignored to monosemantic at the top of the phase diagram between points A and B. When features are sparse, we see a smoother transition from ignored to polysemantic to monosemantic representations.

At the critical point (labelled C in the figure), the marginal loss benefit curves are flat and equal in scale, so \emph{any} allocation of capacity that makes full use of the available embedding dimensions is optimal. Thus, any given feature could be ignored, represented monosemantically, or represented polysemantically.

The shapes of the $\partial L/\partial C_i$ curves are specific to our toy model, but it is generally true that the boundaries between phases are determined by the marginal loss benefit of capacity.
Note that this is true for \emph{any} notion of capacity satisfying the properties discussed in Section~\ref{sec:loss_min}.

Finally, we note that our phase diagram bears a striking resemblance to that of a physical system with a second-order phase transition: there is a line of discontinuous ``first order'' change (above the critical point), giving way to a continuous transition at higher sparsity. The resemblance is stronger if we think of $\partial L/\partial C_i$ as our ``order parameter" instead of $C_i$ itself: $\partial L/\partial C_i$ is constant along one side of the discontinuity, and changes as a function of sparsity along the other side, with the discontinuity going to zero at the critical point. $\partial L/\partial C_i$ is continuous everywhere else. That said, there are some notable differences from typical second-order transitions; most importantly, $\partial L/\partial C_i$ is not a smooth analytic function of importance and sparsity, but rather a piecewise-analytic function whose derivatives are discontinuous at the edges of the superposition phase. It is also unclear whether there is any meaningful analog of critical-point phenomena such as critical exponents.

\subsection{Other models and nonlinearities}

We now compare the numerical phase diagram for our toy model to that of Anthropic's autoencoder model~\cite{XYZ}.
The key difference between these models is that we sum the output of the nonlinear layer before computing the loss whereas in Anthropic's model the loss is computed by summing the squared differences on the output layer.
This means that theirs is tasked with data compression (i.e. recovering an encoded vector) while ours is tasked with a nonlinear computation.

We compare these models across three nonlinearities: quadratic, ReLU, and GeLU.
In our model when we vary the nonlinearity we do so both in the ground truth and in the activations.
We further consider two cases: one with six features embedded in three dimensions (``6 in 3'') and one with six features embedded in five dimensions (``6 in 5'').

\begin{figure*}
    \centering
    \includegraphics[width=1.0\textwidth]{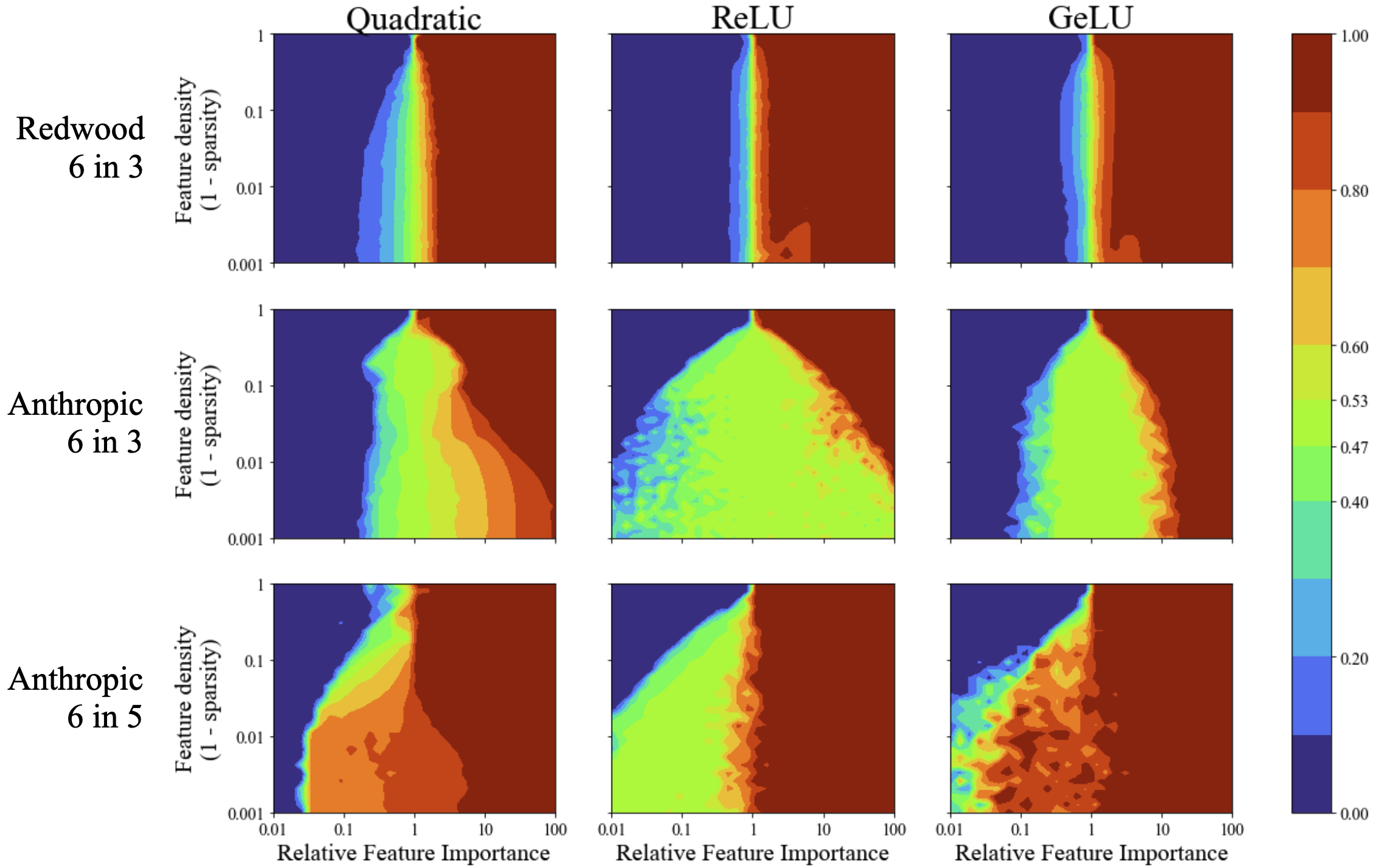}
    \caption{Empirical phase diagrams are shown for our toy model, Anthropic's toy model with 6 features in 3 dimensions, and Anthropic's toy model with 6 features in 5 dimensions. For each model we study three different nonlinearities. Our toy model looks similar for 6 features in 3 dimensions and 6 features in 5 dimensions for all nonlinearities, so we only display plots of 6 features in 3 dimensions. The contour at $c=0.5$ has been replaced with two contours at $c=0.47,0.53$ to give more detail for the large $c\approx 1/2$ regions, which correspond to the antipode polytope from ~\cite{XYZ}.}
    \label{fig:phase-grid}
\end{figure*}

The top row of Figure~\ref{fig:phase-grid} shows the results for our toy model. The same general shape is the same in all cases, as predicted from our theoretical results: a sharp 0-1 transition at mild sparsity (low kurtosis), and a smoother transition with a region of superposition for more extreme sparsity (higher kurtosis).

The superposition phase of our model also seems to be smoother, with capacity changing steadily from 0 to 1 (as predicted by our analytic results). In contrast, the Anthropic model has a large region where the capacity used by a feature is ``stuck'' close to 1/2. This presumably corresponds to an antipodal embedding geometry; or, in the terminology of Section \ref{sec:matrices}, a separate block of the embedding matrix containing two features in a one-dimensional subspace. If the neuron basis is rotated appropriately, this becomes a single neuron that represents two features, with neither the neuron nor the features overlapping with anything else. By contrast, the superposed features in our model unavoidably project onto many neurons regardless of how the neuron basis is rotated.

Interestingly, some aspects of the phase diagram seem to depend on the nonlinearity and/or model. Our model has a narrower superposed region than Anthropic's model. For the ``6 in 3'' case with significantly more features than neurons, Anthropic's model shows features in superposition even when their importance is orders of magnitude larger than any other feature, which does not happen for our model (or for Anthropic's when there is only one more feature than neurons).

\section{Geometry of efficient matrices}\label{sec:matrices}

In this section we study the geometric of \emph{efficient} embedding matrices, namely those which saturate the total capacity bound $\sum_i C_i \leq D$ with the capacity defined in equation~\eqref{eq:capacity}.
In particular, we exhaustively enumerate the forms these matrices can take and comment on the implications for the optimal capacity allocation in models.

\subsection{Diagonal (monosemantic)}

The simplest way to make $W$ efficient is to put $D$ of the vectors orthogonal to each other with arbitrary lengths and set the rest to zero. This makes $W$ a diagonal matrix padded with zeros. 

In this case, the capacities $C_i$ are either 0 or 1.
The norms-squared $n_i$ are zero (if $C_i=0$) or nonzero but arbitrary (if $C_i=1$).

\subsection{Semiorthogonal (``everything bagel'')}

At the other extreme, we can choose $W=\sqrt{\lambda} R$ where $R$ is semiorthogonal, meaning that $RR^T=I$. (This means the rows of $W$ are orthonormal, but the columns aren't.) 

In this case, we have $WW^T=\lambda I$, so
\begin{align}
(W^TW)^2 &=\lambda W^TW
\end{align}
and the capacity simplifies to
\begin{align}
C_i = \frac1\lambda [W^TW]_{ii} = \frac{n_i}\lambda
\end{align}
So the capacities and norms of vectors are directly related.

We now confirm that this is efficient:
\begin{align}
\sum_i C_i &= \frac1\lambda\text{Tr}(W^TW)= \frac1\lambda\text{Tr}(WW^T)= \frac1\lambda\lambda D= D
\end{align}

\begin{thm}
Such a matrix exists for \emph{any} $\vec C$ that sums to $D$, so this gives us a way of dividing up capacity efficiently however we like.
\end{thm}
The proof is shown in Appendix~\ref{sec:joe}.

The singular values of a semiorthogonal matrix are all equal (and vice versa), so the condition of semiorthogonality is equivalent to requiring embedding vectors to be distributed in all directions in a balanced, isotropic way.\footnote{The exact sense of ``balanced" is a little tricky: the vectors don't have to be centered around the origin! Instead, think about a dataset whose principal components all have the same variance.} This gives some geometric intuition for what the angles between vectors are doing: shorter vectors, for features with less capacity, will be spaced closer together in order to keep the overall arrangement balanced.

\subsection{Mixed (tegum product)}

We can combine these approaches by splitting our embedding space into orthogonal subspaces of dimension $D_k$ and using a semiorthogonal matrix to embed $N_k$ of our vectors into the $k$th subspace.

To do this, make $W$ block-diagonal with blocks $\lambda_kR_k$, where $R_k$ is a $D_k\times N_k$ semiorthogonal matrix.
Here $\sum_k D_k = D$ and $\sum_k N_k = N$.
For an example with two blocks, see Figure \ref{fig:embedding-geometry}.

Within each block, the dimensions of a subspace are divided up as before:
\begin{align}
n_i &= \lambda_k C_i \quad \text{if $i \in$ block }k\\
\sum_{i \in \text{ block }k}C_i&=D_k
\end{align}
Note that the independent scalars $\lambda_k$ let us scale different subspaces arbitrarily. (This is the only thing that makes this more general than the previous case.)

\subsection{General}

So far we have chosen subspaces aligned with the standard basis on $\mathbb R^D$ (i.e. the neurons), but we can rotate our set of embedding vectors however we like without changing $\vec C$ or $\vec n$.
Similarly, we have assigned dimensions of $\mathbb R^N$ (i.e. features) to subspaces sequentially, but we can do it in any order.

In other words: a sufficient condition for efficiency is that $W$ is a $D\times N$ matrix of the form $QBP$, where
\begin{itemize}
\item $Q$ is a $D \times D$ orthogonal matrix,
\item $P$ is an $N \times N$ permutation matrix, and
\item $B$ is a $D\times N$ block-diagonal matrix whose (rectangular) blocks are of the form $R_k \sqrt{\gamma_k}$, where $\gamma_k$ is a positive scalar and $R_k$ is a semiorthogonal matrix.
\end{itemize}
In terms of the SVD decomposition $W = QSR$, an equivalent condition is that rows of $R$ corresponding to unequal singular values $S_{ii}$ are never nonzero in the same column.

\begin{thm}
This is also a necessary condition; all efficient matrices are of this form.
\end{thm}
The proof is shown in Appendix~\ref{sec:joe}.

\subsection{Interpolating between regimes}

Consider some block of an efficient matrix (factor of a tegum product), and some vector $\vec w_i$ in that block. If the norm-squared $n_i$ of that vector is increased, its capacity $C_i$ increases in order to keep $c\propto n$ within the block. At the same time, the vector becomes more orthogonal to the rest of the block. 

If the capacity reaches $C_i=1$, the vector becomes fully orthogonal and separates into a new block. Once that happens, $n_i$ can be varied arbitrarily without affecting $C_i$.

On the other hand, if the capacity decreases to $C_i=0$, then $n_i=0$ and the vector becomes zero; at that point the feature is ignored.

\subsection{Application to toy models}

Empirically, Anthropic found that (when importances are equal) their embedding matrices often factor into orthogonal subspaces containing simple geometric structures~\cite{XYZ}. This is equivalent to the statement that their embedding matrices have a large number of small blocks. (This also explains some features of the phase diagram for their model, such as a large region near $c=1/2$; see above.) In contrast, the embedding matrices in our model have one large superposed block (except for ignored or fully-represented features).

We can understand this phenomenon by thinking about the constraints on $\vec C$ and $\vec n$ in the large-block and small-block regime. When there is one large block, capacity can be divided up arbitrarily; however, the lengths of embedding vectors are determined entirely by the capacities (up to an overall scalar). For our model, this is compatible with good loss, because minimizing loss gives the relationship $C_i \propto n_i$ anyway (see above). However, for Anthropic's model, this may be more constraining of embedding vector norms than the loss function ``wants".

In contrast, a collection of small blocks imposes constraints on capacity allocation, because each block must separately add up to the correct capacity. Once capacity is allocated, however, the lengths of vectors can be chosen much more freely by scaling blocks independently of each other. This doesn't provide any benefit for our toy model, but seems to be useful to Anthropic's.

This qualitative difference matters because it affects the interpretability of the network: when there are many small blocks, it is possible to choose an orthogonal basis of neurons such that each neuron is only polysemantic between a small number of features. On the other hand, large blocks imply that neurons will be polysemantic between many features at once. 

\section{Conclusions}

We have studied polysemanticity through the lens of feature \emph{capacity}, or the fraction of an embedding dimension allocated to each feature.
Treating capacity allocation as a constrained optimization problem, we find that many features are either ignored (not embedded) or else allocated a full embedding dimension orthogonal to all other features, depending on their relative importance to the loss.
Features are represented polysemantically only when the marginal loss reduction of assigning more capacity to each is equal and significant (Figure \ref{fig:deriv-curves}).
This neatly explains the sharp ``pinning'' of features to either 0 or 1 capacity noted by \cite{XYZ}.

To explore capacity allocation in a concrete model, we investigated a 2-layer model with quadratic activations (Section~\ref{sec:quadratic}) and constructed a phase diagram of capacity allocation.
We found good qualitative agreement to the phase diagram of \cite{XYZ}, and confirmed our theoretical predictions regarding optimal capacity allocation.

Finally, we studied the geometry of the embedding space (Section~\ref{sec:matrices}), and characterized embedding matrices which make full use of the available dimensions.
We found that efficient embedding matrices are block-semiorthogonal.
These blocks correspond to the polytope structure \cite{XYZ} found, with small blocks corresponding to features embedded as regular polytopes and large blocks corresponding to less-ordered structures.
Large- and small-block arrangements come with different advantages.
With large blocks there is significant freedom to allocate capacity across features, whereas with small blocks there is the additional constraint that the capacity of each block be an integer and that the block capacities add up to the total capacity.
On the other hand, with small blocks the lengths of embedding vectors can be chosen more freely because blocks can be scaled independently of each other without affecting the capacity allocation.

In our quadratic model the embedding matrices in our toy model always have one large block, while \cite{XYZ} found a range of block sizes depending on the properties of the training distribution.
This suggests that differences in architecture can lead to different sizes of blocks, which could provide a way to control the extent of polysemanticity in models, alongside other approaches such as changing the activation function~\cite{elhage2022solu}.

\section*{Author Contributions}

AS derived the geometry of efficient matrices, proved the upper bound on the model capacity, and proposed an alternate definition of capacity.
KS performed the numerical calculations in this work.
ASJ, KS, and BS developed the capacity allocation and constrained optimization framework.
AS, KS, and BS developed the quadratic model, and derived the expected loss, and related this to the input kurtosis and sparsity.
AS, KS, ASJ, and BS derived the analytic optima for the quadratic model.
JB proved the existence of feasible matrices for optimal capacity allocations and proved that single-block semiorthogonal matrices can be found for any capacity allocation. 
JS, KS, ASJ, and BS provided feedback on drafts and the organization of this manuscript.
AS, ASJ, and KS contributed to writing this manuscript.

\section*{Acknowledgments}
The Flatiron Institute is supported by the Simons Foundation.
We are grateful to Jacob Steinhardt, David Lindner, Oliver Balfour, Jeff Wu, Ryan Greenblatt, and Martin Wattenberg for helpful comments on this manuscript, as well as to Chris Olah and Nicholas Schiefer for discussions about this work.
We thank Chris Olah for producing Figure~\ref{fig:schema2}.

\appendix

\section{Quadratic Model Expected Loss}\label{appen:loss}

Our loss function is squared error
\begin{align}
L(x) = (\tilde y(x) - y(x))^2.
\end{align}
We  assume that the $x_i$'s are IID with $\mathbb{E}[x_i]=0$ and $\text{Var}[x_i]=1$. With this, we can write the expected loss as
\begin{align}
\mathbb{E}[L]&=\mathbb{E}[(x^T Dx + b)^2]\\
&= \mathbb{E} \left[ \left( \sum_{i,j} D_{ij}x_ix_j +b\right) ^2 \right] \\
&= \sum_{i,j,k,l}D_{ij}D_{kl} \mathbb{E} \left[ x_ix_jx_kx_l \right]+2b\sum_{i,j}D_{ij}\mathbb{E}[x_ix_j]+b^2,
\end{align}
where $D=W^T W - v$ and $v$ is the diagonal matrix formed of $\{v_i\}$.

We can split the first summation term into several summations by considering all partitions of four terms (4, 3|1, 2|2, 2|1|1, 1|1|1|1). For example, the 2|2 cases are $\sum_{(i=j)\neq(k=l)}, \sum_{(i=k)\neq(j=l)}, \sum_{(i=l)\neq(k=j)}$. Because the $x$'s are independent with mean 0 and variance 1, the only partition terms that are non-zero are 4 and 2|2, so we find
\begin{align}
\mathbb{E}[L] &= \mathbb{E}[x_i^4]\sum_i D_{ii}^2 + \sum_{i \neq j}D_{ii}D_{jj}+2\sum_{i \neq j}D_{ij}^2+2b\sum_i D_{ii}^2+b^2 \\
&= (\mathbb{E}[x_i^4]-1)\sum_i D_{ii}^2 + \sum_{i,j}D_{ii}D_{jj}+2\sum_{i \neq j}D_{ij}^2+2b\sum_i D_{ii}^2+b^2 \\
&= (\mathbb{E}[x_i^4]-1)\sum_i D_{ii}^2 + \left(\sum_{i}D_{ii}\right)^2+2\sum_{i \neq j}D_{ij}^2+2b\sum_i D_{ii}^2+b^2 \\
&= (\mathbb{E}[x_i^4]-1)\sum_i D_{ii}^2 + \left(b + \sum_{i}D_{ii}\right)^2+2\sum_{i \neq j}D_{ij}^2.
\end{align}

Inspecting the second term we see that at optimum $b=-\sum_i D_{ii}$.
Making this assignment we find
\begin{align}
\mathbb{E}[L]&= (\mathbb{E}[x_i^4]-1)\sum_i D_{ii}^2 +2\sum_{i \neq j}D_{ij}^2,
\end{align}
Substituting in for $D$ gives us
\begin{equation}
\mathbb{E}[L]= (\mathbb{E}[x_i^4]-1)\sum_i (||W_{\cdot,i}||^2-v_i)^2 +2\sum_{i \neq j}(W_{\cdot,i} \cdot W_{\cdot,j})^2 
\end{equation}

\section{Covariances and Correlations}\label{appen:covar}

Here we derive the relationship between the terms in the quadratic model loss and true/hallucinated correlations:
\begin{align}
\text{Cov}[y,x_i^2] &= \text{Cov}\left[\sum_j v_j x_j^2,x_i^2 \right] \\
&=v_i\text{Cov}[x_i^2,x_i^2] \\
&=v_i(\mathbb{E}[x_i^4]-\mathbb{E}[x_i^2]^2) \\
&=v_i(\mathbb{E}[x_i^4]-1) \\
\text{Cov}[\tilde{y},x_i^2] &= \text{Cov}\left[\sum_{j,k} (W^TW)_{jk}x_jx_k,x_i^2\right] \\
&= (W^TW)_{ii}\text{Cov}[x_i^2,x_i^2] \\
&= ||W_{\cdot,i}||^2(\mathbb{E}[x_i^4]-1) \\
\\
\text{Cov}[y,x_ix_j] &= 0 &(i \neq j) \\
\text{Cov}[\tilde{y},x_ix_j] &= 
\text{Cov}\left[\sum_{j,k} (W^TW)_{jk}x_jx_k,x_ix_j\right] &(i \neq j)\\
&= 2(W^TW)_{ij}\text{Cov}[x_ix_j,x_ix_j] \\
&= 2(W_{\cdot,i} \cdot W_{\cdot,j}) \\
\end{align}
Here the true correlations in the model are described by the term $\text{Cov}[\tilde{y},x_i^2]$, as these are what appear in the ground truth ($\text{Cov}[y,x_i^2]$).
By contrast, the hallucinated correlations are those of the form $\text{Cov}[\tilde{y},x_ix_j], i\neq j$, which are zero in the ground truth but are generally non-zero in the quadratic toy model.

\section{Quadratic Model Loss Minmization}\label{appen:minimize}

In minimizing the loss we have to consider three constraints:
\begin{enumerate}
    \item $1 \le \sum_i C_i \le D$
    \item $0 \le C_i \le 1$ 
    \item $(\vec C, \vec n)$ is feasible
\end{enumerate}
Here feasibility means that there is actually a set of embedding directions we can choose such that $W$ realizes the capacity allocation $\vec C$ and embedding lengths $\vec n$. 
Our strategy here will be to ``ask forgiveness rather than permission": We minimize $\mathcal L$, ignoring the feasibility constraint, and check for feasibility after the fact.

First we hold $\vec C$ fixed and optimize over $\vec n$:
\def\ka{(k-1)}
\begin{align}
\frac{\partial \mathcal{L}}{\partial{n_i}} &= 2\ka(n_i-v_i)-4n_i+4\frac{n_i}{C_i}\\
0&=2\ka (n_i^*-v_i)-4n_i^*+4\frac{n_i^*}{C_i}\\
n_i^* ((k-3)+2/C_i)&=\ka v_i\\
n_i^*&=\frac{(k-1)v_i}{(k-3)+2/C_i}
\end{align}
Note that $\lim_{C_i\to 0} n_i^*=0$ and $\lim_{C_i\to0} (n_i^*{}^2/C_i)=0$, so the final two terms of the loss are zero for non-represented features. 

Next we differentiate with respect to capacity to find
\begin{align}
\frac{\partial\mathcal L(\vec C, \vec n^*(\vec C))}{\partial C_i} = \frac{\partial \mathcal L}{\partial C_i} + \sum_j\frac{\partial \mathcal L}{\partial n_j}\frac{\partial n_j^*}{\partial C_i}
\end{align}
The second term vanishes\footnote{This is because we optimized over $\{n_i\}$ without any constraints, so $\vec n^*$ is a local optimum for any $\vec C$.}, so we only have the first term
\begin{align}
\def\ka{\kappa_4}
\frac{\partial \mathcal L}{\partial C_i} = -2\frac{(n_i^*(C_i))^2}{C_i^2}
\end{align}
This is non-positive, so our loss is monotonically decreasing in $C_i$. That means that we will saturate our first constraint, $\sum_i C_i = D$.

Next let's solve for $C_i$ and $n_i$:
\begin{align}
\lambda C_i &= \frac{(k-1)v_i}{(k-3)+2/C_i}\\
\frac{v_i}\lambda &= C_i\frac{k-3}{k-1} + \frac2{k-1}\\
\label{eq:Ci_alloc}
C_i &= \max\left(\min\left(\frac{k-1}{k-3}\frac{v_i}{\lambda} - \frac2{k-3}, 1\right), 0\right)
\end{align}
where we have introduced the clipping at 0 and 1 to make this also apply to the capacities on the boundary.
With $C_i$ we then obtain
\begin{align}
n_i = \begin{cases}
0 &\quad C_i=0\\
\frac{k-1}{k-3}v_i-\frac{2\lambda}{k-3} &\quad 0< C_i < 1\\
v_i&\quad C_i=1
\end{cases}
\end{align}

\subsubsection{Checking feasibility}

For our solution $(\vec C, \vec n)$ to be feasible, there must exist a corresponding matrix $W$.

As shown in Appendix~\ref{sec:joe}, for any $\vec C$ with $\sum_i C_i = D \le N$ and $0\le C_i\le 1$, there is a semiorthogonal matrix $W$ of shape $(N,D)$ with $C_i(W) = C_i$.

We next need to show that $\vec n$ is compatible with $\vec C$. The condition $n_i \propto C_i$ holds for any semiorthogonal matrix, and we can multiply each $W_{\cdot,i}$ by a scalar to get the constant of proportionality right.

Therefore, this solution is feasible!

\section{Kurtosis and Sparsity}\label{sec:misc}

\subsection{Sparsity increases fourth moment}

Suppose that, with probability $p$, $x$ takes the value of a new random variable $z$ and otherwise it is 0 (i.e. $x=sz$, $s \sim B(p)$). We set the mean $\mathbb{E}[z]=0$ and variance $\text{Var}[z]=1/p$ so that $\mathbb{E}[x]=0, \text{Var}[x]=1$. As $p$ decreases (i.e. sparsity increases), $\mathbb{E}[x^4]$ increases:

\begin{align}
    \mathbb{E}[x^4]=\mathbb{E}[s^4z^4]=\mathbb{E}[s^4]\mathbb{E}[z^4]=p\cdot\text{Kurt}[z]\text{Var}[z]^2=\text{Kurt}[z]/p
\end{align}

\subsection{Uniform Random Variable Moments}

We want to choose $a$ such that for $x \sim U(-a,a) \cdot B(p)$, $\text{Var}[x]=1$. Choose $a=\sqrt{\frac{3}{p}}$:

\begin{align}
\text{Var}[x]&=\mathbb{E}[s^2]\mathbb{E}[z^2]=p\cdot\frac{\left(2\sqrt{3/p}\right)^2}{12}=1 \\
\text{Kurt}[x]&=\frac{\mathbb{E}[x^4]}{\mathbb{E}[x^2]^2}=\frac{9}{5}\cdot\frac{1}{p}
\end{align}

\section{Interpretation as correlation coefficient}\label{app:rho2}

The (population) Pearson correlation coefficient $\rho^2$, also called ``fraction of variance explained'', is defined as the ratio between covariance squared and the product of variances:

\begin{align}
\rho^2(X, Y) = \frac{\ka(X, Y)^2}{\kappa(X, X)\kappa(Y, Y)}
\end{align}

(The more-familiar $r^2$ is an estimator for $\rho^2$ computed from a sample.)

Let's assume our input features $X_i$ are linearly uncorrelated and have mean zero and variance one:

\begin{align}
\ka(X_i) &= 0\\
\ka(X_i, X_j) &= \delta_{ij}
\end{align}

We'll embed them with embedding vectors $w_i = W\hat e_i = W_{:,i}$ and immediately unembed with $W^T$. Then $\rho^2$ between the $i$th input and output is (using Einstein notation for everything except $i$, which is never summed)

\begin{align}
\rho^2(X_i, [W^TWX]_i) &= \frac{\ka(X_i, W*{ik}W_{kj}X_j)^2}{\ka(X_i, X_i)\ka(W_{ik}W_{kj}X_j, W_{ik'}W_{k'j'}X_{j'})}\\
&= \frac{(W_{ik}W_{kj}\ka(X_i, X_j))^2}{W_{ik}W_{kj}W_{ik'}W_{k'j'}\ka(X_i, X_i)\ka(X_j, X_{j'})}\\
&= \frac{(W_{ik}W_{kj}\delta_{ij})^2}{W_{ik}W_{kj}W_{ik'}W_{k'j'}\delta_{jj'}}\\
&= \frac{(W_{ik}W_{ki})^2}{(W_{ik}W_{kj})(W_{ik'}W_{k'j})}\\
&= \frac{(w_i\cdot w_i)^2}{\sum_j (w_i \cdot w_j)^2}\\
&= C_i
\end{align}

where we've written the sum over $j$ explicitly for the sake of clarity.

Note that $W^T$ is not always the optimal way to reconstruct $X_i$ from $WX$. We haven't checked the math, but it seems to be the case that using the optimal unembedding matrix instead of $W^T$ gives the alternate definition of capacity in Appendix \ref{app:alt-capacity}.

\section{Proof of capacity constraint}\label{app:cap-constraint}

Suppose we have $N$ features and $D$ neurons, with $N \ge D$.\\

Let our embedding matrix be $W_{ai}$, with shape $[D, N]$. The embedding vectors in $\mathbb R^D$ are column vectors $\vec{w}_i := W_{:,i}$ with components $[w_i]_a = W_{ai}$.

Define
\begin{align}
V_{ai} &:= \frac{W_{ai}}{\sqrt{[W^TWW^TW]_{ii}}}\\
\vec v_i := V_{:,i} &= \frac{\vec w_i}{\sqrt{\sum_j (\vec w_i \cdot \vec w_j)^2}}
\end{align}
and
\begin{align}
U &:= V^T W \\
U_{ij} &= \vec v_i \cdot \vec w_j
\end{align}
Note that
\begin{align}
U_{ii}^2 &= \frac{(\vec w_i \cdot \vec w_i)^2}{\sum_j (\vec w_i \cdot \vec w_j)^2} = C_i
\end{align}
Also,
\begin{align}
\sum_{j=1}^N U_{ij}^2 = 1
\end{align}
so the row vectors $\hat u_i := U_{i,:} \in \mathbb R^N$ are unit vectors.\\

Let $D' := \text{rank}(U) \le D$, so that the vectors $\hat u_i$ span a subspace of dimension $D'$.\\

Let the $D'$ unit vectors $\hat y_b\in \mathbb R^N$ with components $[y_b]_i =: Y_{bi}$ be an orthonormal basis for this subspace.\\

Let $X_{ib}$ be the components of $\hat u_i$ in the $\hat y_b$ basis, so that
\begin{align}
U &= X Y\\
\hat u_i &= \sum_{b=1}^{D'} X_{ib} \hat y_b
\end{align}
$\hat u_i$ is a unit vector, so its components in an orthonormal basis obey
\begin{align}
\sum_{b=1}^{D'} X_{ib}^2 &= 1
\end{align}
Therefore, the row vectors $\hat{x}_i := X_{i,:} \in \mathbb R^{D'}$ are unit vectors.\\

We can also define vectors $\vec{y}_i := Y_{:,i}$; note that these are the columns of $Y$ and $\hat y_b$ are the rows.\\

Note that $U_{ij} = \hat x_i \cdot \vec y_i$, so we have
\begin{align}
C_i = U_{ii}^2 &= (\hat x_i \cdot \vec y_i)^2\\
& \le (\hat x_i\cdot \hat x_i) (\vec y_i\cdot \vec y_i) \text{\quad by Cauchy-Schwarz}\\
&= (\vec y_i\cdot \vec y_i)\\
&= \sum_{b=1}^{D'} Y_{bi}^2
\end{align}

Finally, summing over $i$,
\begin{align}
\sum_{i=1}^N C_i &\le \sum_{i=1}^N\sum_{b=1}^{D'} Y_{bi}^2\\
&= \sum_{b=1}^{D'} \sum_{i=1}^N Y_{bi}^2\\
&= \sum_{b=1}^{D'} |\hat y_b|^2\\
&= D' \le D
\end{align}

\section{An alternative definition of capacity}\label{app:alt-capacity}

Instead of applying equation (2) directly to $W$, we can take the (compact) SVD $W=QSR$ and apply it to $R$,

\begin{align}
\tilde C_i(W) :=\frac{([R^TR]_{ii})^2}{[(R^TR)^2]_{ii}}=[R^TR]_{ii}
\end{align}

An equivalent prescription is to first rotate (by $Q^{-1}$) and then scale (by $S^{-1}$) the embedding space $\mathbb R^D$ so that the embedding vectors are isotropically distributed, then take $\tilde C_i$ to be the norm-squared of the new embedding vector $S^{-1}Q^{-1}\vec w_i$.

This is equivalent to our previous definition when $W$ is ``$c$-efficient'',

\begin{align}
\sum_{i=1}^N C_i=D\implies \tilde C_i = C_i
\end{align}

but relaxes the condition for ``$\tilde c$-efficiency'' dramatically, to the point that generic matrices are efficient:

\begin{align}
\sum_{i=1}^N\tilde C_i &= \text{rank}(W)\\
W\text{ full rank}&\implies \sum_{i=1}^N \tilde C_i=D
\end{align}

On the other hand, $\tilde C_i(W)$ is a somewhat more discontinuous function of $W$ than $C_i(W)$ is.

Our results for the quadratic toy model are essentially unchanged; $W$ is efficient at the minimum of $\mathbb \mathbb{E}[\mathcal L]$, so these definitions coincide (although we're not sure their derivatives do). In more realistic models, it remains to be seen which version is more useful.

This new definition also sheds some light on ``$c$-efficiency'': a matrix is efficient if its embedding vectors are isotropically distributed within each block. In the case of only one block, a matrix is efficient if identity covariance for a random $\vec x$ implies identity covariance for $W\vec x$.

\section{Feasibility Proofs}\label{sec:joe}

\paragraph{Set-up:} We have vectors $w_1, \dots, w_N \in \R^D$ spanning $\R^D$, where $D \leq N$, arranged into an $D \times N$ matrix
\begin{equation*}
    W = \begin{pmatrix}
    | & & | \\
    w_1 & \dots & w_N \\
    | & & |
    \end{pmatrix}.
\end{equation*}
Our aim is to show that
\begin{equation*}
    \sum_{i=1}^N C_i \leq D
\end{equation*}
and to determine the conditions under which equality holds.

\paragraph{Solution:} We begin by finding a maximal decomposition of $\R^D$ into orthogonal subspaces $V_1 \oplus \dots \oplus V_r$ such that each $w_i$ lies in some $V_s$ and partition the $w_i$ accordingly, discarding any index with $w_i=0$. It then suffices to examine the problem separately on each subspace $V_s$.

Therefore, we assume that $w_1, \dots, w_N$ are such that we cannot decompose our problem further. In particular, this implies that $w_i \neq 0$ for all $i$ and we cannot partition the indices into two subsets $S,T$ such that $w_i \cdot w_j = 0$ for all $i \in S, j \in T$.

\begin{lemma}
Let $\{e_1, \dots, e_N\}$ be the standard basis of $\R^N$ and let $V \subseteq \R^N$ be a linear subspace of dimension $D$, such that $e_i \not \in V^\perp$ for all $i$. Then for any arbitrary unit vectors $v_1, \dots, v_N \in V$, we have
\begin{equation*}
    \sum_{i=1}^N (v_i \cdot e_i)^2 \leq D
\end{equation*}
with equality if and only if $v_i$ is parallel to $\proj(e_i, V)$, where $\proj(u, W)$ denotes the orthogonal projection of a vector $u$ onto the subspace $W$.
\end{lemma}

\emph{Proof:}
Define $v_i^\ast \in V$ to be a unit vector in the direction $\proj(e_i, V)$, and note that this exists by the assumption $e_i \not \in V^\perp$. Then we have $|v_i \cdot e_i| \leq |v^\ast_i \cdot e_i|$ for all other unit vectors $v \in V$, with equality if and only if $v_i$ is parallel to $v^\ast_i$. So it suffices to show that
\begin{equation*}
    \sum_{i=1}^N (v_i^\ast \cdot e_i)^2 = D.
\end{equation*}
Let $b_1, \dots, b_D$ be an orthonormal basis for $V$, and let $b_{D+1}, \dots, b_N$ be an orthonormal basis for $V^\perp$. Then \begin{equation*}
    v_i^\ast = \lambda_i (b_1 \cdot e_i) b_1 + \dots + (b_D \cdot e_i) b_D
\end{equation*}
where the normalising constant $\lambda_i = (\sum_{j=1}^D (b_j \cdot e_i)^2)^{-1/2}$ is non-zero as $v^\ast_i$ is non-zero (and we may wlog assume $\lambda_i > 0$). Then
\begin{equation*}
    (v_i^\ast \cdot e_i)^2 = \lambda_i^2 \lr{\sum_{j=1}^D (b_j \cdot e_i)^2 }^2 = \sum_{j=1}^D (b_j \cdot e_i)^2
\end{equation*}
from which the lemma follows.

Define the matrix $A = (a_{ij})_{i,j=1, \dots, N}$ via $a_{ij} = w_i \cdot w_j$, and denote by $a_i$ the vector $(a_{i1}, \dots, a_{in})^T \in \R^N$, so that
\begin{equation*}
    A = \begin{pmatrix}
    | & & | \\
    a_1 & \dots & a_N \\
    | & & |
    \end{pmatrix}
\end{equation*}
and the capacities can be expressed as $C_i = a_{ii}^2 / ||a_i||^2$. Note that $a_{ii} = ||w_i||^2 \neq 0$ and so non of the $a_i$ are trivial. Also, note that $A = W^T W$, so the rank of $A$ is the same as the rank of $W$, and thus equal to $D$.

Now let $v_i = a_i / ||a_i||$, so that $v_i$ is a unit vector. Since the rank of $A$ is $D$, the $a_i$ span a space $V \subseteq \R^N$ of dimension $D$. We can write
\begin{equation*}
    v_i \cdot e_i = a_{ii} / ||a_i||,
\end{equation*}
from which it also follows that $e_i \not \in V^\perp$ (otherwise we would have $a_{ii} = 0$ which is not possible). Therefore we may apply Lemma 1 to get
\begin{equation*}
    \sum_{i=1}^N C_i \leq D
\end{equation*}
with equality if and only if $a_i$ is parallel to $\proj(e_i, V)$.

Assuming that equality holds, using the proof of Lemma 1, we can write
\begin{equation*}
    a_i = \mu_i \sum_{k=1}^D (b_k \cdot e_i)b_k, \hspace{5mm}
    a_{ij} = \mu_i \sum_{k=1}^D (b_k \cdot e_i) (b_k \cdot e_j)
\end{equation*}
for some constant $\mu_i$. Since $a_{ii} = a_i \cdot e_i > 0$, we see that $\mu_i > 0$.

Now, if $\mu_i \neq \mu_j$ then
\begin{equation*}
    \mu_j a_{ij} = \mu_j \mu_i \sum_{k=1}^D (b_k \cdot e_i)(b_k \cdot e_j) = \mu_i a_{ji}
\end{equation*}
from which we conclude by the symmetry of $a_{ij}$ that $a_{ij} = w_i \cdot w_j = 0$. Hence if we have any $\mu_i \neq \mu_j$ we can perform a non-trivial decomposition into orthogonal subspaces $V_1 \oplus \dots \oplus V_r$ satisfying the conditions at the start of the solution. Since we assumed this was not possible, we must in fact have $\mu := \mu_i$ is constant for all $i$.

Consider the map
\begin{equation*}
    \Lambda : w_i \mapsto \mu^{1/2} \sum_{k=1}^D (b_k \cdot e_i) b_k.
\end{equation*}
If we let $T \subseteq \R^N$ be the linear subspace spanned by $w_1, \dots, w_N$, then $\dim T = m$ and $\Lambda$ maps $T$ to $V$. We can also check that the orthogonality of the bases $\{e_i\}_i$ and $\{b_i\}_i$ implies that $\Lambda$ preserves the inner products of the $w_i$. It follows that $\Lambda$ is invertible and orthogonal.

In addition, we have that
\begin{equation*}
    b_j^T \Lambda W e_i = b_j^T \Lambda w_i = \mu^{1/2} \sum_{k=1}^D (b_k \cdot e_i)(b_k \cdot b_j) = \mu^{1/2} (b_j \cdot e_i)
\end{equation*}
so if we consider $\Lambda W$ as a map from $\R^D$ with basis $\{e_i\}_i$ to $V$ with basis $\{b_i\}_i$ then it has matrix
\begin{equation*}
    \Lambda W = \mu^{1/2} \begin{pmatrix} b_1 \cdot e_1 & \dots & b_1 \cdot e_N \\
    \vdots & & \vdots \\
    b_D \cdot e_1 & \dots & b_D \cdot e_N
    \end{pmatrix}.
\end{equation*}
Using $\sum_{j=1}^N (b_i \cdot e_j)^2 = 1$, it follows that $\Lambda W W^T \Lambda^T = \mu I_D$. Since $\Lambda$ is orthogonal, we deduce that $WW^T = \mu I_D$ as required.

\paragraph{Showing that every saturating case is achievable:} We want to show that for every tuple of real numbers $(C_1, \dots, C_N)$ with $0 \leq C_i \leq 1$ and $\sum_{i=1}^N C_i = D$, we can find $w_1, \dots, w_N \in \R^D$ such that \begin{equation*}
    C_i = \frac{(w_i \cdot w_i)^2}{\sum_{j=1}^N (w_i \cdot w_j)^2}.
\end{equation*}
A tuple $(C_1, \dots, C_N)$ is feasible if there exists an $N \times N$ unitary matrix $U$ such that if we let $\hat{U}$ denote the first $D$ rows of $U$ and define $u_1, \dots, u_N \in \R^D$ by
\begin{equation*}
    \hat{U} = \begin{pmatrix}
    | & & | \\
    u_1 & \dots & u_N \\
    | & & |
    \end{pmatrix},
\end{equation*}
then $||u_i||^2 = C_i$ for each $i = 1, \dots, n$. Note that if $(C_1, \dots, C_N)$ is feasible then for the corresponding $u_i$, if we denote their components by $\{u_{ki}\}_{k=1}^D$, we have
\begin{align*}
    \frac{(u_i \cdot u_i)^2}{\sum_{j=1}^N (u_i \cdot u_j)^2} & = \frac{(u_i \cdot u_i)^2}{\sum_{j=1}^N\sum_{k,l = 1}^D u_{ik} u_{jk} u_{il} u_{jl}} \\
    & = \frac{(u_i \cdot u_i)^2}{\sum_{k,l = 1}^D u_{ki} u_{li} \delta_{kl}} \\
    & = \frac{(u_i \cdot u_i)^2}{\sum_{k=1}^D u_{ki}u_{ki} } = ||u_i||^2 = C_i
\end{align*}
so we see that $(C_1, \dots, C_N)$ is a valid set of capacities by taking $w_i = u_i$ for each $i$. Hence it suffices to show that all the relevant tuples are feasible.

\begin{lemma}
Suppose that $(C_1, \dots, C_N)$ is a feasible tuple. Then
\begin{enumerate}
    \item any permutation of $(C_1, \dots, C_N)$ is feasible;
    \item for any $C'_N, C'_{N+1} \geq 0$ such that $C'_N + C'_{N+1} = C_N$, the tuple $(C_1, \dots, C_{n-1}, C'_N, C'_{N+1})$ is feasible;
    \item the tuple $(1 - C_1, \dots, 1 - C_N)$ is feasible, with $D$ replaced by $N-D$.
\end{enumerate}
\end{lemma}

\emph{Proof:} Let $U$ be an $N \times N$ unitary matrix corresponding to $(C_1, \dots, C_N)$ as above. Then (i) is immediate by permuting the columns of $U$, and (iii) follows by exchanging the first $D$ and last $N-D$ rows of $U$, noting that the squared norm of each column is 1.

To prove (ii), consider extending $U$ to an $(N + 1) \times (N + 1)$ unitary matrix $U'$ by placing $U$ in the upper $n \times n$ quadrant and placing a 1 at the bottom right position. Then write
\begin{equation*}
    U' = \begin{pmatrix}
  \begin{matrix}
  & & \\
  & U & \\
  & &
  \end{matrix}
  & \rvline & 0 \\
\hline
  0 & \rvline &
  \begin{matrix}
  1
  \end{matrix}
\end{pmatrix}
= \begin{pmatrix}
    | & & | & | \\
    u'_1 & \dots & u'_N & u'_{N+1} \\
    | & & | & |
    \end{pmatrix},
\end{equation*}
for vectors $u'_i \in \R^{N+1}$. Now we may rotate the final pair of vectors together by an arbitrary angle $\theta$ and the resulting matrix remains orthogonal:
\begin{equation*}
    U'_\theta = \begin{pmatrix}
     & & & \\
    u'_1 & \dots & u'_N \cos(\theta) + u'_{N+1}\sin(\theta) & -u'_N \sin(\theta) + u'_{N+1}\cos(\theta) \\
    & & &
    \end{pmatrix},
\end{equation*}
Restricting to the first $D$ rows of $U'_\theta$, we see that the first $n-1$ columns still have squared norms $C_1, \dots, C_{n-1}$ respectively (inherited from $U$). Therefore, as $\theta$ varies the sums of the squared norms of the final two columns must be $C_N$. Moreover, the split between the two columns varies continuously between $(C_N, 0)$ and $(0, C_N)$ as $\theta$ ranges between $0$ and $\pi/2$. Therefore we can achieve any split $(C'_N, c'_{N+1})$ between the final two columns, proving part (2).

Finally, we show that every tuple $(C_1, \dots, C_N)$ satisfying $0 \leq C_i \leq 1$ and $\sum_{i=1}^N C_i = m$ is feasible. Suppose not, for a contradiction. Then we may pick such a tuple with minimal $n$. Clearly all such tuples with $n \leq 1$ are feasible, so we may assume $n \geq 2$.

If there are any $C_i, C_j$ with $C_i + C_j \leq 1$ then we may replace $C_i, C_j$ by $C_i + C_j$ to get another tuple satisfying the given conditions but with smaller $n$. By our minimality assumption, we conclude this new tuple must be feasible. But then our original tuple is feasible by Lemma 2.1 and 2.2.

On the other hand, if there are no $C_i, C_j$ with $C_i + C_j \leq 1$ then consider the tuple $(1 - C_1, \dots, 1 - C_N)$. We know that we must have $(1 - C_i) + (1 - C_j) \leq 1$ for any $i, j$. Replacing $(1 - C_i), (1 - C_j)$ by $(1 - C_i) + (1 - C_j)$ to get a new tuple and following the argument of the previous paragraph, we conclude that the new tuple must be feasible, and so $(1 - C_1, \dots, 1 - C_N)$ must be feasible also. But then $(C_1, \dots, C_N)$ is feasible too by Lemma 2.3. We are thus done in all cases.

\clearpage

\bibliographystyle{halpha}
\bibliography{main}

\end{document}